%% file: main.tex
\documentclass[10pt,conference,letterpaper]{IEEEtran}
\usepackage{times,amsmath,epsfig}
\usepackage{url,subcaption}
\title{Compliance Change Tracking in Business Process Services}
\author{Srikanth G Tamilselvam, Ankush Gupta, Arvind Agarwal \\ IBM Research India
}

\begin{document}
\maketitle
\begin{abstract} 
Regulatory compliance is an organization's adherence to laws, regulations, guidelines and specifications relevant to its business. Compliance officers responsible for maintaining adherence constantly struggle to keep up with the large amount of changes in regulatory requirements. 
Keeping up with the changes entail two main tasks: fetching the regulatory announcements that actually contain \textit{changes of interest}, and incorporating those changes in the business process. In this paper we focus on the first task, and present a Compliance Change Tracking System, that gathers regulatory  announcements from government sites, news sites, email subscriptions; classifies their importance i.e Actionability through a hierarchical classifier, and business process applicability through a multi-class classifier. For these classifiers, we experiment with several approaches such as vanilla classification method (e.g. Na\"ive Bayes, logistic regression etc.), hierarchical classification method, rule based approach, hybrid approach with various preprocessing and feature selection methods; and show that despite the richness of other models, a simple hierarchical classification with bag-of-words features works the best for Actionability classifier and multi-class logistic regression works the best for Applicability classifier. The system has been deployed in global delivery centers, and has received positive feedback from payroll compliance officers.
\end{abstract}

% NOTE keywords are not used for conference papers so do not populate them
% \begin{keywords}
% keyword-1, keyword-2, keyword-3
% \end{keywords}
%
\input{introduction.tex}
\input{litsurvey.tex}
\input{approach.tex}

\input{data_collection.tex}
\input{classification.tex}

\input{ui.tex}
%\input{challenges.tex}
\input{conclusion.tex}

\bibliographystyle{IEEEtran}

\bibliography{IEEEabrv,IEEEexample}

\end{document}

%% file: introduction.tex
\section{Introduction}
\label{sec:intro}
Organizations are faced with rapidly changing regulatory policies, and ever-growing number of regulations. It is estimated that by the year 2020, just the global banks will be required to comply with over 120,000 pages of regulations\footnote{https://www.infosys.com/industries/financial-services/Documents/regulatory-compliance-management.pdf}. Failure in adhering to these regulations often leads to huge monetary fines, customer dissatisfaction, and damage to the reputation of the business; and therefore, Chief Financial Officers see this as their topmost challenge~\cite{cfo}.
% In addition to the volume of regulatory documents, the compliance task is further complicated by the complex language used in these documents~\cite{breaux2005mining} which forces banks to hire domain experts, whose primary job is to identify relevant changes, and accordingly introduce or suggest changes to specific internal controls to remain compliant.

The large amount of regulatory changes published, and the uncertainty about the time and place where they are published, make compliance officers job tremendously challenging, especially because of the risk, non-compliance carries with it. Because of this, organizations are increasingly relying on cognitive technologies to help them with regulatory change tracking. Regulatory Change Tracking helps organizations keep track of changes in regulations, and answer questions like {\it ``what are the changes in the newly published regulation that we care about?"}.
A conventional approach to solve this problem is to deploy a large workforce which constantly keeps track of sources where regulatory bodies publish  changes.
%  Figure~\ref{fig:current_process} captures the current process, where the domain experts in the role of regional compliance officers process paid regulation subscriptions or manually monitor relevant sources for relevant policy additions or changes. An example of such a change is {\it ``Increases the maximum wage base from \$45,252 to \$46,694.''} which is a payroll related change and may need to reflect in organization Payroll, HR systems. 
During such monitoring, if a \textit{relevant} change is found, it is communicated to compliance officers who then perform a gap analysis i.e., compare the changed policy with internal controls supported by external vendors like SAP. Based on the gaps identified, compliance officers recommend and follow up to ensure compliance with the changed policy. This whole process require highly skilled labor that needs to be continuously trained on new regulations. Such a demand of skilled labor has led to growing operational costs in recent years, sometimes accounting for more than 10\% \cite{kpmg} of the total operational expenses. 
 %To address the volume, velocity, variety, and complexity of regulations, banks are increasingly seeking technological help.

%\begin{figure}
%	\center
%	\includegraphics[width=\linewidth]{current_process.png}
%	\caption{\small Current Regulatory Compliance process}
%	\label{fig:current_process}
%\end{figure}

% Employees on a regular basis browse through these sources to first identify {\it relevant} regulations, and compare them with the previous version to identify the changes. This whole manual process is not only time consuming but also prone to errors.

%\begin{figure}
%	\center
%	\includegraphics[width=\linewidth]{new_process.png}
%	\caption{\small Semi Automated Regulatory Compliance process}
%	\label{fig:new_process}
%\end{figure}

In this paper, we propose a cognitive system for tracking changes in regulatory documents pertaining to payroll business process. Our cognitive compliance change tracking system monitors sources where regulatory bodies publish announcements related to regulatory changes and makes the job of the compliance officers easier by categorizing the announcements into appropriate categories according to their actionability and business process applicability. Thus significantly reducing the manual effort spent in reading through the changes to determine their importance and relevance to the organization.
Our contribution in this paper is as follows: 
\begin{itemize}
\item We present a novel application in Compliance Change Tracking, where we process news articles containing regulatory changes, categorize them into appropriate categories,  and present them to compliance officers through an easily consumable user interface.
\item Given the nature of the problem where there are relatively few news articles which are of interest to the user, we experiment with several approaches for categorization including several preprocessing and feature selection methods, and show that a simple hierarchical classification method based on bag-of-words features works the best for Actionability Classifier, and multi-class logistic regression classifier for Applicability Classifier.
\item We demonstrate a use-case of the proposed system in \textit{Payroll business process}, however the system is equally applicable to any other business processes where compliance officers need to track regulatory changes to ensure compliance.
%\item We present an extension of the application for other resource scarce languages. More specifically, we build classifiers for new languages where the supervised data is limited. We provide a transfer learning framework where we transfer the learning used in the existing language to new languages.
\end{itemize}

\begin{figure*}
\centering
\begin{subfigure}{0.49\textwidth}
\centering
\includegraphics[width=\linewidth]{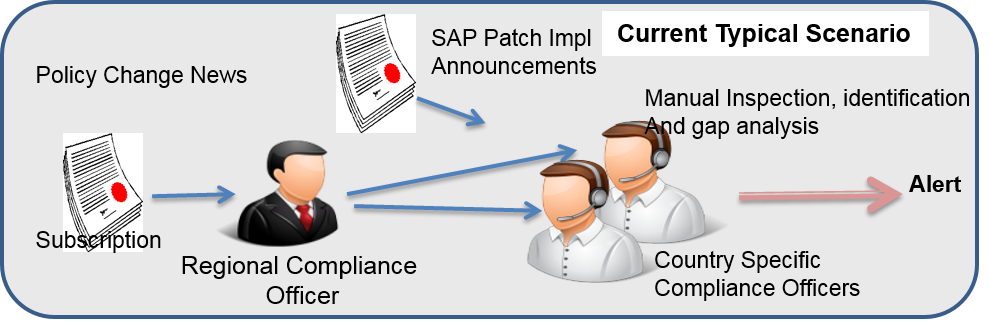}
\caption{\small Current Compliance Change Tracking System}
\label{fig:current_process}
\end{subfigure}
\begin{subfigure}{0.49\textwidth}
\centering
\includegraphics[width=\linewidth]{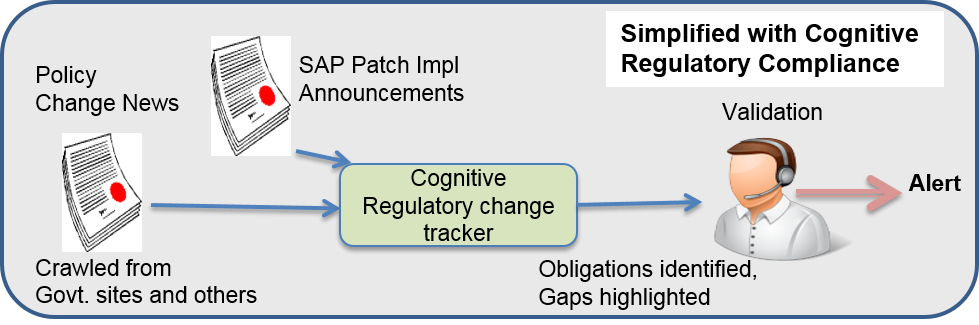}
\caption{\small Cognitive Compliance Change Tracking System}
\label{fig:new_process}
\end{subfigure}
\caption{Current (left) and Proposed (right) Compliance Change Tracking System}
\label{fig:combined}
\end{figure*}

%% file: litsurvey.tex
\section{Background Work in Regulatory Compliance and Change Tracking}
\label{sec:lit}
There are several aspects of regulatory compliance where researchers have used machine learning and data mining to help with the compliance process. Research had been around semantic parameterization for extracting and prioritizing rights and obligations from regulations\cite{breaux2005analyzing, breaux2006towards}. Similar work from WYNER and PETERS\cite{wyner2011rule} came up with extraction rules from regulations which can be used to represent obligations. Both of these works provide a semantic representation of obligations which in turn can be converted into business process execution logic. There has been some work around mapping the obligations with business process\cite{sapkota2016automating} to identify the affected business process in case of any change in the regulation. Despite the importance of the problem\cite{english2012cost, hammond2013cost}, there has been limited work in compliance change tracking, in fact, to the best of our knowledge, there is no academic work particularly in this area. There are some academic work around building ontology for regulatory change management\cite{espinoza2014ontology}, a critical component of enterprise, governance, risk and compliance (EGRC) framework\cite{caldwell2008magic}. Some of the commercial products available in the space of regulatory change management are PREDICT360\footnote{\url{http://www.360factors.com/regulatory-change-management-software/}} from 360Factors, MetricStream\footnote{\url{http://www.metricstream.com/apps/regulatory-change-management.htm}}, Thomson Reuters\footnote{\url{https://risk.thomsonreuters.com/en/resources/infographic/regulatory-change-management.html}}, KPMG\footnote{\url{https://advisory.kpmg.us/risk-consulting/frm/regulatory-change-management-transformation.html}}. These systems are often built on top of domain knowledge collected over the years and with the help of large workforce that constantly keep track of regulatory changes and make them available to subscribers.

%\begin{figure}
%\includegraphics[width=\linewidth]{ria.png}
%  \caption{A screenshot of news articles as available in  subscription service RIA Checkpoint provided by Thomson Reuters}
%  \label{fig:ria}
%\end{figure}

\section{Current Compliance Change Tracking System}
\label{sec:current_process}
In this section, we describe the current compliance change tracking process that compliance officers follow in order to keep their system up-to-date. The very first step in compliance change tracking process is to know about the latest news articles that contain information about the changes that may need to be implemented in the back-end system.  An example of such a change is {\it ``Increases the maximum wage base from \$45,252 to \$46,694.''} which is a payroll related change and may need to reflect in organization Payroll, HR systems. In developed countries such as USA, there are paid professional services available which one can subscribe to get these news articles on a regular basis. These services deploy a large workforce to go through websites that publish such changes. Those changes are then compiled into one place and made available for a fee. One such paid subscription service is RIA by Thomson Reuters\footnote{https://checkpoint.riag.com/}. In other less developed countries, where such services are not available, or those organizations that cannot afford the fee, they seek the help of compliance officers to manually browse through \textit{important} websites and keep track of the changes. Figure~\ref{fig:current_process} captures the current process, where domain experts in the role of regional compliance officers process paid regulation subscriptions or manually monitor relevant sources for policy additions or changes.

Typically, compliance officers go through the news articles,  categorize them (Actionability classifier) into one of three classes (1) \textit{InformationOnly} : News articles that are good-to-know but does not contain any changes to be implemented into the back-end system. (2) \textit{ActionRequired} : News articles that require some changes to be implemented into the back-end system (3) \textit{Irrelevant} : News articles that are not at all relevant for their purpose. In addition to this, compliance officers also classify (Applicability classifier) articles into different topics to ease their job for change tracking and reporting such as Payroll, HR, Benefits etc. It might be worth noting that in the case of paid service subscription, even though all the news articles are available at one place, compliance officers often tend to open the source website to go through the original article, primarily because subscription service often only contain the snippet of the article which may not be enough to understand the context to recommend and implement changes. %Figure~\ref{fig:ria} shows the screenshot of such a paid subscription service RIA by Thomson Reuters containing the snippets of news articles.  organized according to jurisdictions. In RIA, while "Pennsylvania — Unemployment"  snippet captures the key changes, the source article\footnote{\url{http://www.uc.pa.gov/Documents/E-Payment\%20Announcement.pdf}} provides more context with more detailed information.

Besides identifying the Actionability and Applicability of changes in the article, compliance officers also need to extract other information like effective date, applicable jurisdiction etc. which may not be captured by the discussed services.
Once compliance officers collect the exact changes that either require some changes in the internal controls, or requires other agents' attention, they broadcast via available channels. The \textit{ActionRequired} news articles are transformed into software requirement and sent to the IT team while \textit{InformationOnly} articles are sent across for awareness. They help  IT employees and compliance officers to anticipate for potential actionable requirement in near future.

%\begin{figure}
%  \includegraphics[width=\linewidth]{penn.png}
%  \caption{An example of new article as available from original source website.}
%  \label{fig:penn}
%\end{figure}

\section{Cognitive Compliance Change Tracking System}
\label{sec:new_process}
We propose a cognitive system for tracking changes in regulatory documents. Figure~\ref{fig:new_process} captures our mostly automated process wherein the data collection module which can be customized per client, periodically collects and processes relevant news from pre-determined client interest sources such as government websites, news websites, software patch documents released by commercial products, email subscriptions, etc. These documents are then converted into common text format to be passed on to machine learning module for classification. More specifically, we design a two-step hierarchical classifier (Actionability classifier), where first step classifies the documents into relevant vs irrelevant, while the second step classifies relevant document further into \textit{ActionRequired} and \textit{InformationOnly} classes. 
%Figure~\ref{fig:article} contains an example of a document from the \textit{ActionRequired} category where the text snippet containing the policy change is highlighted. 
We further design another classifier (Applicability classifier) which classifies the documents according to the applicable business process. With the help of these two classifiers, compliance officers can readily work on documents or category of interest and focus on more intensive tasks like gap analysis with respect to current operations and estimate the impact to the business due to the changes.

%\begin{figure}
%	\center
%\includegraphics[width=\linewidth]{article.png}
%	\caption{\small An example of an article showing the snippet containing the change.}
%	\label{fig:article}
%\end{figure}

%% file: approach.tex
\section{System Detail}
\begin{figure}
	\includegraphics[width=\linewidth]{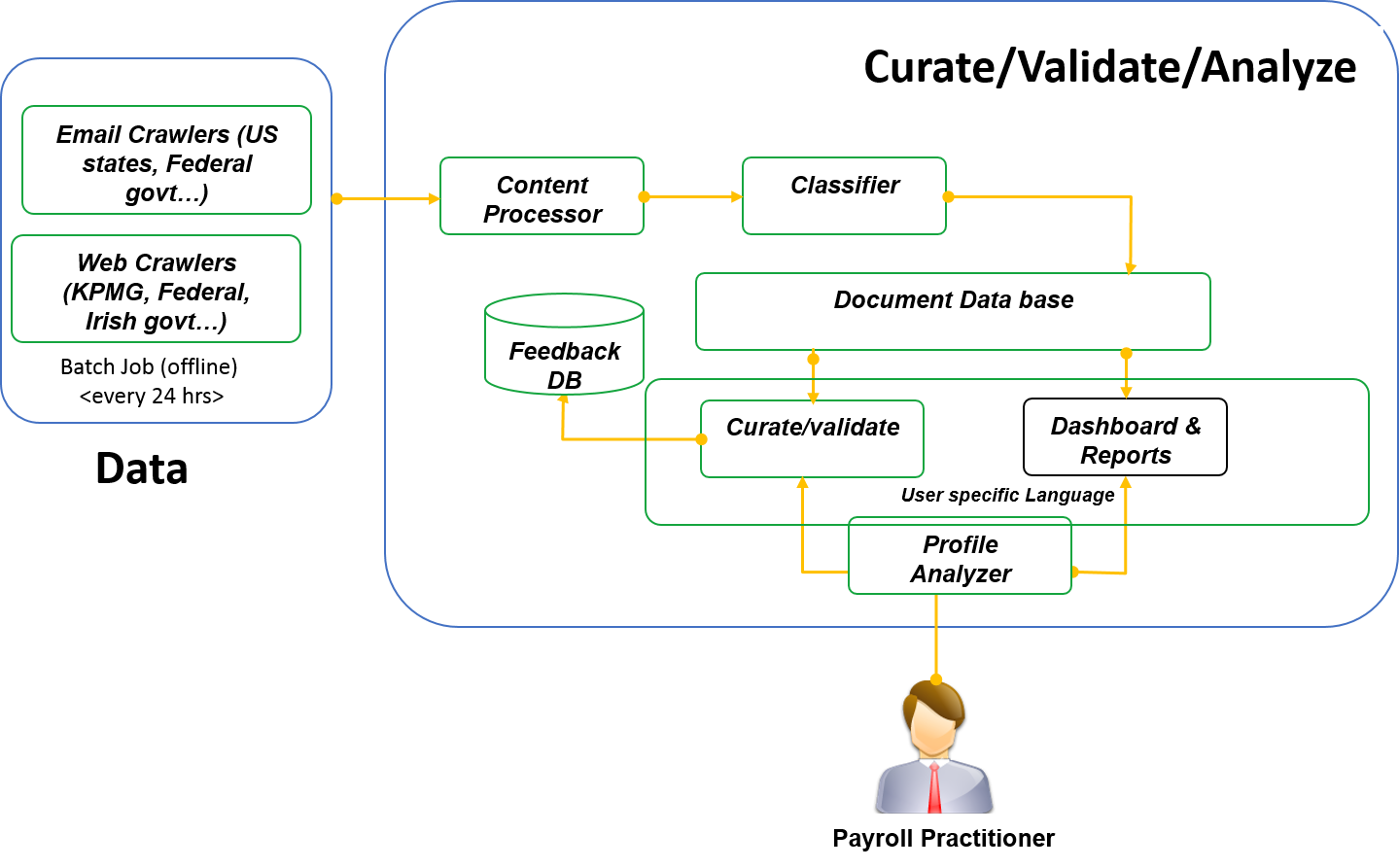}
	\caption{Payroll Compliance System}
	\label{fig:arch}
\end{figure}

The system follows MVC architecture as shown in Figure~\ref{fig:arch}. It consists of mainly three modules (1) Data collection and ingestion module; (2) Classification module; and (3) User Interface. The following sections explains them in detail.

%% file: data_collection.tex
\subsection{Data Collection and Ingestion}
\label{sec:data_collection}
Government, in general publishes regulatory changes as announcements on their web pages, or push them through email subscriptions to public.
% as seen in Figure~\ref{fig:payroll_updates}. 
 There is no single repository or website for these regulatory changes, they are rather available on department or jurisdiction specific web pages; and it is the responsibility of the compliance officers to keep track of these web pages and emails on a daily basis for any relevant updates. 
%In addition to the compliance officers, organizations also deploy lawyers to keep track of court cases that can have any legal binding on the regulations in near future.
%\begin{figure}
%	\includegraphics[width=\linewidth]{updates.png}
%	\caption{Govt. announcements}
%	\label{fig:payroll_updates}
%\end{figure}
%In our work, we worked with compliance subject matter experts to identify websites or subscribe that in the past provided regulatory change announcements.

In our implementation, we worked with compliance Subject Matter Experts (SME's) to identify websites and subscribe to government channels which in the past provided regulatory change announcements.
For email subscriptions, we leveraged Gmail API Service\footnote{https://developers.google.com/gmail/api/} to read email threads. We then built a custom scrapper using Jsoup\footnote{https://jsoup.org/} to download content from these identified websites or subscribed emails on a daily basis. Having a scrapper that work across different websites and emails is a major challenge. In addition to the complexity in handling variety of the sources, some websites don't specify the announcement date or mention updates on already published article, which made distinguishing between already processed announcements to fresh one harder.
%The downloaded content contain payroll and additional information which brings the necessity to classify content which is explained in later section. Another challenge in the data gathering is the data source and type. 
To overcome these challenges, we customized our scraper to check the length of the announcement(html/pdf) content along with the date to identify if there is any new addition/updation to the website since previous crawl.

We tracked  regulatory change announcements across North America (USA), South America, parts of Asia, Africa, and Europe. We focused on the announcements specific to Payroll process. In North America, out of $50$ states, $31$ states provided payroll announcements on their web pages while $28$ states provided through subscriptions. $9$ states provided through both. In addition to the state specific announcements,Federal announcements are also of interest. These announcements are available through $3$ email subscription channels and $1$ website. In South America, we tracked announcements for $11$ countries, for each country, $4$ government websites namely taxes, labor ministry, social security, national official publications are tracked for announcements. Similarly, relevant sites for Europe (UK, Ireland), and Asia region (India) are also tracked.
%and ignore other parts of email like from, to, email sent datetime etc. 
In total, we crawled, $2721$ government announcements. Out of which $1706$ is from North American state and federal announcements, Africa $433$, India $178$, Europe $97$, South America $307$.

%% file: classification.tex
\subsection{Classification}
\label{sec:classification}
As explained in Section~\ref{sec:current_process}, one of the key requirements of the system is to identify which articles require action from the compliance officer, which articles are only for information, and which ones are irrelevant. We therefore classify articles into these $3$ categories (1) \textit{ActionRequired} (2) \textit{InformationOnly}, and (3) \textit{Irrelevant}. Another requirement from the system is to categorize the articles into appropriate business  processes so that compliance officers can filter which are (1) \textit{Benefits}, (2) \textit{Expats}, (3) \textit{HR}, (4) \textit{Payroll}, (5) \textit{TaxFiling} and (6) \textit{Others}.

\begin{table}
\centering
\begin{tabular}{|c|c|p{1.5cm}|p{1.5cm}|c|}
\hline
Data & Total & Action Required & Information Only & Irrelevant \\
\hline
Historical Data & 420 & 100 & 256 & 64 \\
SME Data 		& 432 & 32 & 117 & 283 \\
Total 			& 852 & 132 & 373 & 347 \\
\hline
\end{tabular}
\caption{Statistics on Historical data available from SME's past analysis, and data available from SME's annotation}
\label{tab:hist-sme-data}
\end{table}

\begin{table}
\centering
\begin{tabular}{|c|c|p{2.0cm}|p{2.0cm}|c|}
\hline
Data & Total & Action Required & Information Only & Irrelevant \\
\hline
Train & 722 & 122 & 338 & 262 \\
Test & 130 & 10 & 35 & 85 \\
Total & 852 & 132 & 373 & 347 \\
\hline
\end{tabular}
\caption{Actionability Classifier Data Statistics: Train on Historical(100\%) + SME(70\%) \& Test on SME data(30\%)}
\label{tab:news-train-test-data-stats}
\end{table}

\begin{table*}[htb]
\centering
\begin{tabular}{|c|c|p{2.8cm}|p{2cm}|c|c|c|}
\hline
Method & Accuracy & Action Required (P/R/F) & Information Only & Relevant & Irrelevant & Average \\
	%& 	& (P/R/F) & (P/R/F) & (P/R/F) & (P/R/F) & (P/R/F) \\
\hline

LR & \textbf{71\%} & \textbf{.60/.60/.60} & \textbf{.50}/.43/.46 & \textbf{.65}/.58/.61 & .79/.84/\textbf{.81} & \textbf{.63/.62/.62}\\
NB & 63\% & 0/0/0 & .38/.51/.43 & .56/.60/.58 & .78/.75/.77 & .39/.42/.40\\
SVM & 40\% & 0/0/0 & .29/.83/.43 & .39/\textbf{.87}/.53 & .79/.27/.40 & .36/.37/.28\\
AdaBoost & 62\% & 0/0/0 & .38/.37/.38 & .51/.42/.46 & .72/.79/.75 & .37/.39/.38\\
RandomForrest & 65\% & .50/.10/.17 & .44/.34/.39 & .43/.27/.33 & .71/.85/.77 & .55/.43/.44\\
KNeighbors & 48\% & .29/.20/.24 & .34/\textbf{.86/.49} & .41/\textbf{.87/.56} & \textbf{.83}/.35/.50 & .49/.47/.41\\
ExtraTrees & 65\% & .33/.10/.15 & .46/.34/.39 & .47/.16/.23 & .71/\textbf{.85}/.77 & .50/.43/.44\\
DecisionTree & 60\% & .13/.10/.11 & .38/.43/.40 & .48/.53/.51 & .76/.73/.74 & .42/.42/.42\\
BaggingClassifier & 65\% & .5/.1/.17 & .4/ .34/.37 & .55/.40/.46 & .72/.84/.78 & .54/.43/.44\\
\hline
\end{tabular}
\caption{Actionability Classifier results for 3-class classification. \textit{Relevant} class results are computed by combining \textit{ActionRequired} and \textit{InformationOnly}. The numbers in a cell are in the order of Precision/Recall/F-score}
\label{tab:news_results_3_class}
\end{table*}

\begin{table*}[htb]
\centering
\begin{tabular}{|c|c|p{2.8cm}|p{2cm}|c|c|c|}
\hline
Method & Accuracy & Action Required (P/R/F) & Information Only & Relevant & Irrelevant & Average \\
	%& 	& (P/R/F) & (P/R/F) & (P/R/F) & (P/R/F) & (P/R/F) \\
\hline
LR & \textbf{71\%} & .50/\textbf{.70/0.58} & \textbf{.50}/.51/\textbf{.50} & .64/.71/\textbf{.67} & \textbf{.84}/.79/\textbf{.81} & 0.61/\textbf{0.67/0.63}\\
NB & 66\% & \textbf{.67}/.20/0.30 & .41/.54/.47 & .59/.64/.62 & .80/.76/.78 & \textbf{0.63}/0.50/0.52\\
SVM & 33\% & 0/0/0 & .27/\textbf{.89}/.41 & .36/\textbf{.91}/.52 & .75/.14/.24 & 0.34/0.34/0.22\\
AdaBoost & 67\% & .33/.30/.31 & .46/.46/.46 & .61/.60/.61 & .79/.80/.79 & 0.53/0.52/0.52 \\
KNeighbors & 47\% & .40/.20/.26 & .33/.83/.47 & .41/.84/.55 & .81/.35/.49 & 0.51/0.46/0.41 \\
ExtraTrees & 66\% & .50/.10/.17 & .43/.29/.35 & .60/.33/.43 & .71/\textbf{.88}/.79 & 0.55/0.42/0.44 \\
DecisionTree & 62\% & .13/.10/.11 & .39/.51/.44 & .56/.67/.61 & .80/.72/.76 & 0.44/0.44/0.44 \\
BaggingClassifier & 68\% & .33/.20/.25 & .46/.37/.41 & \textbf{.65}/.49/.56 & .76/.86/\textbf{.81} & 0.52/0.48/0.49 \\
\hline
\end{tabular}
\caption{Actionability Classifier results for hierarchical classification. The numbers in a cell are in the order of P/R/F-score}
\label{tab:news_results_hierarchy}
\end{table*}

%We had access to the subject matter experts (SMEs) who could provide us the supervision for the tasks. These SMEs have been doing this analysis for years so they already had some of the data available, however this data was different from real-world data, so we also asked SMEs to annotate real-world data. Since annotation task is time consuming, to make the task easier, we built an annotation tool which can be used by compliance officers to expedite the annotation process. The tool uses an initial classifier to provide some initial labels, and then ask users to mark them as correct or incorrect. In the case of incorrect labels, user is asked to provide the correct label. Thus this tool can be used for both online evaluation as well as for collecting the supervised data. For data collection. we used an initial rule based classifier and asked domain experts to give us the correct data. 
%
%We considered supervised classification based approach for this problem and for any supervised learning annotated dataset becomes the primary importance. 
We employed supervised classification methods for this problem. Our supervised data came from two sources. First source was the historical data where compliance officers had already categorized the articles available from other sources such as RIA into appropriate categories. As mentioned in Section~\ref{sec:lit}, RIA only provides a hand crafted short snippet of the whole article therefore this data was very different from the data that we encounter in real-world setting. In order to have our classifiers learn from the real world data, we added a second source of data, where we asked subject matter experts (SME's) i.e compliance officers to manually label them. Since data annotation is a time consuming task, we built an annotation tool to expedite the annotation process. The tool provides initial labels based out of simple rule based classifier, and then prompts users to mark them as correct or incorrect. In the case of incorrect labels, user is asked to provide the correct label with reasons. This tool can be used for both, online evaluation as well as for collecting the supervised data. 
The statistics of both of these data sources are shown in Table~\ref{tab:hist-sme-data}. As seen from the table, the historical data is more balanced in terms of number of examples in each class. This is because the categorization is done on the articles that are provided from the subscription service, and since subscription services already filters a lot of \textit{Irrelevant} articles, it results in a balanced set. On the other hand, the categorization provided on the SME data, directly sourced from websites, is more unbalanced. We see relatively high number of \textit{Irrelevant} articles compared to other two classes. Since in our final system deployment, the classification will be performed on the articles which are directly sourced from the websites, we choose the test data to be only from SME data. In order to keep the test set constant across the experiments, we take 30\% of the SME data to be the test data, and the rest is used for training.

Now we describe the experiments performed on both classifiers and discuss their results in detail.

\subsubsection{Actionability Classifier}
Among both classifiers, this classifier is more important since it is acceptable to misclassify the business process applicability of an article however it is not acceptable to misclassify the actionability of an article. As long as actionability of an article is correctly identified, it will be taken up by some business unit for action. Furthermore, compliance officers are more concerned with the correctness of the \textit{ActionRequired} class than the other two classes. It is worth noting here that there is a natural hierarchy among these three classes, i.e., \textit{ActionRequired} and \textit{InformationOnly}, both can be merged into one class called \textit{Relevant}. While the compliance officers would like to look at the \textit{ActionRequired} articles since those are the articles that they are concerned with the most, they also look at the \textit{InformationOnly} articles. So it is acceptable to a certain extent to classify \textit{ActionRequired} articles into \textit{InformationOnly} class and vice versa, however, classifying them into \textit{Irrelevant} is not admissible. \textit{Irrelevant} articles are something that the compliance officer would ignore. Given the nature of the problem, we experiment with two settings (1) Treating the classification problem as 3-class problem; (2) Treating it as a hierarchical classification problem where we first classify articles into \textit{Relevant} vs \textit{Irrelevant} classes and then relevant classified articles are further classified into \textit{InformationOnly} and \textit{ActionRequired}.

\begin{table*}[htb]
\centering
\begin{tabular}{|c|c|}
\hline
Class & Rules\\
\hline
\textit{ActionRequired} & withholding, wage, rate, employee, payroll, tax rate, tax table, personal tax exemption, personal exemption, etc.\\
\textit{InformationOnly} & benefit, leave, publish, will not change the resident tax rates, proposed, may pay, subject to approval, etc.\\
\textit{Irrelevant} & license, sales, occupational, drilling rules, cigarette floor tax, corporation business tax, gift or estate tax, hotelroom tax, etc.\\
\hline
\end{tabular}
\caption{Examples of Rules}
\label{example-rules}
\end{table*}

\paragraph{Evaluation Metrics}
Since we are dealing with an imbalance class classification problem, we use Precision, Recall and F-measure as our evaluation metrics. 
%Although we report Precision, Recall, F-score for all three classes, our prime concern are the numbers for \textit{ActionRequired} class. Even for this class, our focus will be on recall since this is what is most important for compliance officers. Recall measures the ratio of the number of articles  correctly classified with the total number of articles available. A low recall for a class would mean that we are missing on the correct classification for that class. 
Missing an \textit{ActionRequired} item carries the highest risk, hence our prime concern is the Recall for this class. Our next focus would be on the ``Relevant" category. While this category is naturally available for hierarchical classifier, it is not available for $3$-class classifier so we artificially create it by combining the \textit{InformationOnly} and \textit{ActionRequired} classes. Similar to the \textit{ActionRequired} class, here also our focus will be on recall.

\paragraph{Results Analysis}
In the first setting where we treat the problem as a 3-class classification problem, we experiment with several Machine learning (ML) algorithms\footnote{We used Scikit-learn package available in Python for building ML models} such as Logistic Regression (LR), Naive Bayes (NB), Support Vector Machines (SVM), KNeighbors and DecisionTree. The statistics of the dataset used for training and testing these classifiers is shown in Table~\ref{tab:news-train-test-data-stats}, and the performance results of different classifiers are shown in Table~\ref{tab:news_results_3_class}.  Our classifiers come from a variety of categories i.e. linear discriminative classifier (LR), non-linear discriminative classifier (SVMs), generative classifier (NB), non-linear classifiers (Decision Tree, KNN), and ensemble classifiers (Random Forest, AdaBoost, Extra Trees, Bagging Classifiers). 
%Recall that although this is a 3-class classification problem, we also report the results for \textit{Relevant} vs. \textit{Irrelevant} problem. The results for \textit{Relevant} vs. \textit{Irrelevant} are computed by simply merging \textit{InformationOnly} and \textit{ActionRequired} in one class i.e. \textit{Relevant} class. 
Despite the richness of other models, Logistic Regression performs the best among all classifiers across different metrics followed by K-Nearest Neighbors. For Logistics Regression, we get the best Recall, Precision and F-score for \textit{ActionRequired} class, and the best Precision and F-score for \textit{Relevant} class. When taking average of all three metrics across all three classes, LR tops all classifiers.

In the second experimental setting where we treat the problem as a hierarchical classification problem, our problem naturally splits into two different classification steps:
\begin{enumerate}
\item{Step1} : Classify into \textit{Relevant} / \textit{Irrelevant}. All the \textit{ActionRequired} / \textit{InformationOnly} articles are marked as \textit{Relevant}.
\item{Step2} : Classify the articles marked as Relevant in Step1 into \textit{ActionRequired} / \textit{InformationOnly}.
\end{enumerate}
The results of this setting are shown in Table~\ref{tab:news_results_hierarchy}. In order to make the comparison easier, the table shows the results for both classifiers, and are in the same format as the table for $3$-class classifier (Table \ref{tab:news_results_3_class}). Note that the performance metrics for Step2 i.e. \textit{InformationOnly} vs. \textit{ActionRequired} Classifier are reported by evaluating them with respect to the original ground truth, and not with respect to only Step2. This means that we take into account the error encountered in Step1 as well while computing the metrics for Step2, so that these results are directly comparable with the results of 3-class classifier. When comparing across different classifiers, the trend is similar to what we see in 3-class classifier. Logistic Regression achieves the best Recall and F-score for the \textit{ActionRequired} class. It also gets the best F-score for \textit{Relevant} class, and the best Recall and F-score for the average across all three classes.

When comparing 3-class and hierarchical classifiers, we see considerable improvement in hierarchical classifier. For the metric that we care about the most, i.e. \textit{ActionRequired} recall, hierarchical classifier achieves 0.70 compared to 0.60 in 3-class classifier, though we lose on precision here. For \textit{Relevant} recall, hierarchical classifier is 13 percent point better than 3-class classifier while losing only 1 percent point in precision. Even when we take average across all 3-classes, we get 5 percent point improvement in recall while losing only 2 percent point in precision. Hierarchical classifier performs better in most of the cases in average F-score if we were to take that as a measure of overall improvement. The results of 3-class classifier are skewed as we can see from \textit{ActionRequired} recall, precision and F-score numbers for NB, SVM and AdaBoost while for hierarchical classifier they are more balanced. Even for our metrics of interest(\textit{ActionRequired} and \textit{Relevant} recall), hierarchical classifier does better in most of the cases.

%\begin{table*}[htb]
\begin{table*}[tb]
\centering
\begin{tabular}{|c|c|p{2cm}|p{2cm}|c|c|}
\hline
Method & Accuracy & Action Required & Information Only & Relevant & Irrelevant \\
\hline
LR & \textbf{68\%} & \textbf{.32/.6/.41} & \textbf{.51/.54/.53} & \textbf{.63/.78/.69} & \textbf{.86/.75/.81} \\
NB & 65\% & .29/.4/.33 & .42/.51/.46 & .6/.76/.67 & .85/.73/.78 \\
SVM & 58\% & .29/.5/.37 & .35/.43/.38 & .52/.69/.59 & .8/.66/.72 \\
AdaBoost & 67\% & .32/.6/.41 & .47/.49/.48 & .62/.76/.68 & .85/.75/.8 \\
\hline
\end{tabular}
\caption{Actionability Classifier results for hierarchical Hybrid (ML + rule-based) classification. The numbers in a cell are in the order of Precision/Recall/F-score.}
\label{tab:hybrid_news_results_hierarchy}
\end{table*}

%\begin{table*}[htb]
\begin{table*}[tb]
\centering
\begin{tabular}{|c|c|c|c|c|c|c|c|}
\hline
Data & Total & Benefits & Expats & HR & Others & Payroll & TaxFiling \\
\hline
Train & 1245 & 77 & 9 & 7 & 90 & 776 & 286 \\
Test & 186 & 11 & 3 & 2 & 39 & 86 & 45 \\
Total & 1431 & 88 & 12 & 9 & 129 & 862 & 331 \\	
\hline
\end{tabular}
\caption{Applicability Classifier Data Statistics: Train on Historical(100\%) + SME data(70\%) and Test on SME data(30\%)}
\label{tab:category-train-test-data-stats}
\end{table*}

%\begin{table*}[htb]
\begin{table*}[]
\centering
\begin{tabular}{|c|c|c|c|c|c|c|c|c|}
\hline
Method & Accuracy & Benefits & Expats & HR & Others & Payroll & TaxFiling & Average\\
\hline
LR & \textbf{60\%} & .87/\textbf{.63}/\textbf{.73} & \textbf{1}/\textbf{.33}/\textbf{.5} & 0/0/0 & \textbf{.65}/\textbf{.38}/\textbf{.48} & \textbf{.61}/.74/\textbf{.67} & .51/\textbf{.55}/\textbf{.53} & \textbf{.61}/\textbf{.44}/\textbf{.49} \\
NB & 53\% & \textbf{1}/.18/.31 & 0/0/0 & 0/0/0 & .37/.28/.32 & .56/.74/.64 & .55/.47/.51 & .41/.28/.29 \\
SVM & 46\% & 0/0/0 & 0/0/0 & 0/0/0 & 0/0/0 & .46/1/.63 & 0/0/0 & .08/.17/.11 \\
RandomForest & 56\% & .75/.55/.63 & 0/0/0 & 0/0/0 & .48/.36/.41 & .55/.83/.66 & .62/.29/.39 & .4/.34/.35 \\
\hline
\end{tabular}
\caption{Applicability Classifier results for 6-class classification. The numbers in a cell are in the order of P/R/F-score}
\label{tab:category_results_6_class}
\end{table*}

One of the reasons for logistic regression (LR) to perform the best among all classifiers is the inherent properties of the data. Logistic Regression is a discriminative classifier, and discriminative classifiers are known to perform better than generative classifiers when there is less training data \cite{ng2002discriminative}. Since our training data set is relatively small, LR performing better than generative  nai\"ve bayes classifier is not surprising. Similarly better performance of LR compared to other discriminative classifier such as SVM can be attributed to the high number of features. LR is a linear classifier while SVM is non-linear. In the case of large number of features, linear classifier is sufficient to represent the data while non-linear will tend to over-fit due to its excessive representational power.  

For all the above experiments, we used default parameters setting for all hyperparameters available, and used unigram and bigram as our features.
%The total number of features thus came out to be 7386. In order to test the classification performance with respect to different features, we experimented with several feature selection methods. In particular, we performed the following Feature Selection techniques: 
%\begin{enumerate}
%\item  Removed \textit{Irrelevant} features like Numbers, String with numbers, String containing words like month names (jan, feb, etc), city names, etc. Feature set reduced to 6397. 
%\item Used Porter Stemmer. Cleaned feature set like in 1. Features reduce to 5506.
%\item Used NLTK WordNet Lemmatizer. Cleaned feature set like in 1.
%\item Consider only finance related terms using a finance dictionary. Features reduce to 3884.
%\end{enumerate}
%However, none of the above experiments resulted in any improvement 
We experimented with different feature selection experiments but none of them resulted in any improvement 
in overall results. We also experimented with Rule-based classification by creating some rules from the training data. Some of the rules we used are listed in Table V. We also filtered the rules with high precision (greater than 0.5 threshold) and created a hybrid classifier. The results for hierarchical Hybrid classification (ML + Rule-based) are shown in Table \ref{tab:hybrid_news_results_hierarchy} which shows no improvement in performance metrics. We also experimented with different thresholds settings for both \textit{Relevant} and \textit{Irrelevant} classes, but did not find any significant improvement in performance.

%\begin{table*}
%\centering
%\begin{tabular}{|c|c|c|c|c|c|c|c|}
%\hline
%Data & Total & Benefits & Expats & HR & Others & Payroll & TaxFiling \\
%\hline
%Historical Data & 815 & 53 & 1 & 3 & 0 & 577 & 181 \\
%SME Data 	& 616 & 35 & 11 & 6 & 129 & 285 & 150 \\
%Total 		&	1431 & 88 & 12 & 9 & 129 & 862 & 331 \\
%\hline
%\end{tabular}
%\caption{Statistics on Applicability Historical data available from SME's past analysis, and data available from SME's annotation}
%\label{tab:hist-sme-data-applicability}
%\end{table*}

\subsubsection{Applicability Classifier}
Our other classification task is to classify the articles according to the business process they are applicable to. Similar to the Actionability classifier, we experimented with several classifiers on applicability classifier as well. The train-test dataset used for this classifier is shown in Table~\ref{tab:category-train-test-data-stats}. The table shows that unlike the Actionability classifier, the data in applicability classifier is highly unbalanced. This is primarily because we are dealing with the payroll compliance business process. Our train-test split for this classifier is 70-30\%, and was constructed similar to the Actionability classifier. The results of different classifiers are shown in Table~\ref{tab:category_results_6_class}. The relative performance of different classifiers is similar to what we see for Actionability classifier. Logistic Regression performs best among all classifiers. While all other classifiers produce skewed results, classifying articles in few of the classes, Logistic Regression produces more balanced results, resulting in best average precision, recall and F-score across all classes.

%% file: ui.tex
\subsection{User Interface}
Once the news articles are classified into appropriate categories, they are stored in Cloudant database\footnote{https://cloudant.com/} and are made available to the compliance officers through User Interface in the form of a Dashboard. 
% It is hosted in IBM Bluemix\footnote{https://console.bluemix.net/} for external view and can be accessed through \url{http://payrollcompliance.mybluemix.net/}. 
%as shown in figure~\ref{fig:ui} %.
In addition to classification categories, we also extracted other meta information such as effective date, applicable jurisdictions. The Dashboard provides functionality to filter news articles based on geographical region, Actionability category, Applicability category, effective date, etc. To ensure compliance officers only see the data that pertains to their geographical region, we enabled authorization via authentication module. A separate user interface is provided for administrator to add users and create their profile. For further reporting and analysis, an option to search or export data into CSV is also available.

%\begin{figure}
%	\includegraphics[width=\linewidth]{penn.png}
%	\caption{RIA Checkpoint by Thompson Reuters}
%	\label{fig:ria}
%\end{figure}

%\begin{figure}
%	\includegraphics[width=\linewidth]{ui.jpg}
%	\caption{Payroll Dashboard}
%	\label{fig:ui}
%\end{figure}

%% file: conclusion.tex
\section{Conclusion}
In this paper, we presented a Cognitive Compliance Change Tracking System that can monitor online sources where regulatory bodies publish regulatory changes. The system uses machine learning methods to process regulatory changes and classify them into the business process they are applicable to. In addition, among all the changes published, the system identifies the changes which require some action be taken by compliance officers. We have deployed this system inside the organization for payroll officers and is currently under evaluation, however we have already started receiving positive feedback in terms of increase in productivity. 
Despite the reasonable performance of the system, there still remains lots of challenges that need to be addressed in future. Future work will focus on reducing manual interventions in adding new news sources for existing or new regions, improve accuracy in calling out actionable insights, summarize articles by extracting core changes, promote collaboration and knowledge sharing.

%In this paper, we have studied various challenges relating
%to regulatory change tracking. We have then introduced a system
%that addresses many of these challenges through a combination
%human assisted crawler to extract meaningful news articles, text analytics support and machine learning  to process and classify and equally well supported by a interactive user interface. We have deployed the tool inside IBM for payroll officers and achieved significant benefits in terms of productivity improvements.
%Future work will focus on reducing the manual interventions
%in expanding new sources for existing or new regions, improve accuracy in calling out actionable insights, summarize articles in easy consumable way,  promote collaboration and knowledge sharing.
\label{sec:conclusion}